# Extended Mixture of MLP Experts by Hybrid of Conjugate Gradient Method and Modified Cuckoo Search

Hamid Salimi[1], Davar Giveki[1,2], Mohammad Ali Soltanshahi[1], Javad Hatami[1]

[1]School of Mathematics and Computer Science, University of Tehran, Tehran, Iran
`salimi.hamid86@gmail.com, ali.soltanshahi@gmail.com, jvdhtm@gmail.com`
[2] Department of Computer Science Universität des Saarlandes, Saarbrücken, Germany
`s9dagive@stud.uni-saarland.de`

## Abstract

*This paper investigates a new method for improving the learning algorithm of Mixture of Experts (ME) model using a hybrid of Modified Cuckoo Search (MCS) and Conjugate Gradient (CG) as a second order optimization technique. The CG technique is combined with Back-Propagation (BP) algorithm to yield a much more efficient learning algorithm for ME structure. In addition, the experts and gating networks in enhanced model are replaced by CG based Multi-Layer Perceptrons (MLPs) to provide faster and more accurate learning. The CG is considerably depends on initial weights of connections of Artificial Neural Network (ANN), so, a metaheuristic algorithm, the so-called Modified Cuckoo Search is applied in order to select the optimal weights. The performance of proposed method is compared with Gradient Decent Based ME (GDME) and Conjugate Gradient Based ME (CGME) in classification and regression problems. The experimental results show that hybrid MSC and CG based ME (MCS-CGME) has faster convergence and better performance in utilized benchmark data sets.*

## 1. INTRODUCTION

Combining classifier is one of the most popular approaches in pattern recognition, which leads to have a better classification. It increases the recognition rate and improves the reliability of the system. It is usually a good approach in complicated problems due to the small sample size, class overlapping, dimensionality, and substantial noise in the input samples.

Previous experimental and theoretic results show that the combining classifiers with each other lead to higher performance when the base classifiers have small error rates, and their errors are different [1]; in other words, the base classifiers make uncorrelated decision in this case. Generally, classifier selection and classifier fusion are two types of combining classifiers [2]. One of the most popular methods of classifier selection is ME, originally proposed by Jacobs et al. [3]. The ME models the conditional probability density of the target output by mixing the outputs from a set of local experts, each of which separately derives a conditional probability density of the target output. The outputs of expert networks are combined by a gating network which is trained to select the expert(s) that is performing the best at solving the problem [4, 5, and 6]. In the basic form of ME [3], the expert and gating networks are linear classifiers, however, for more complex classification tasks, the expert and gating networks could be of more complicated types. For instance, Ebrahimpour et al. [7] proposes a face detection model, in which they used Multi-Layer Perceptrons (MLPs) [8, 9, 10] to form the gating and expert networks in order to improve the face detection accuracy.

The Back Propagation (BP) algorithm is the most commonly used technique for training neural networks. It is an approximation of the Least Mean Square (LMS) algorithm, which is based on the steepest descent method. While the BP technique follows a straightforward and well-established algorithm, there are some disadvantages associated with it. For instance, the convergence behavior of the BP algorithm highly depends on the choice of initial values of connection weights and other parameters used in the algorithm such as the learning rate and the momentum term [11, 8].

To tackle the mentioned problems, we decided to employ much more powerful methods such as Conjugate Gradient (CG) and Modified Cuckoo Search (MCS) algorithm as a hybrid model. The CG method as a second-order optimization techniques is applied to the learning of MLP in ME structure. The MCS algorithm proposed by Walton et al [12] is an improved version of another metaheuristic algorithm the so called Cuckoo Search (CS) [13] and it can be used for initialing optimal values of connection weights. This is an evolutionary optimization algorithm inspired by the nature.  This algorithm is inspired by the reproduction strategy of cuckoos. At the most basic level, cuckoos lay their eggs in the nests of other host birds, which may be of different species. The host bird may discover that the eggs which are not it's own and either destroy the egg or abandon the nest all together. This has resulted in the evolution of cuckoo eggs which mimic the eggs of local host birds. Experimental results show that this modification causes faster convergence and better performance in classification and regression benchmark problems.

The rest of paper is organized as follow: In Section 2, Conjugate Gradient Multi-Layer Perceptronsis is explained. Mixture of Expert method is introduced in Section 3. Section 4 discusses Cuckoo Search and Modified Cuckoo Search. Experimental results are discussed in Section 5; finally, the conclusion is made in Section 6.

## 2. CONJUGATE GRADIENT MULTI-LAYER PERCEPTRONS (CGMLP)

The BP learning algorithm is a supervised learning method for multi-layer feed-forward neural networks. It is a gradient descent local optimization technique which involves backward error correction of network weights. Despite the general success of BP method in the learning process, several major deficiencies are still needed to be solved. The convergence rate of BP is slow and hence it becomes unsuitable for large problems. Furthermore, the convergence behavior of the BP algorithm depends on the choice of initial values of connection weights and other parameters used in the algorithm such as the learning rate and the momentum term.

One way for speeding up the learning phase is using higher-order optimization. In the BP case the function is approximated by only the linear terms that include the first-order derivatives in a Taylor series expansion. In this case the second-order nonlinear terms that include the second-order derivatives are also used in the Taylor series expansion, resulting more precise approximation [14].

Given a vector $\Delta w_0$ in the weight space, a second-order Taylor series approximation of the error function around this vector is expressed as

$$E(w) = E(w_0) + d^T \Delta w + \frac{1}{2}(\Delta w)^T H \Delta w \qquad (1)$$

where $d, H$ are the gradient vector and the Hessian matrix, respectively. The minima of the function $E$ are located where the gradient of $E$ was equal to zero, therefore, the optimal value of $w$ is given by

$$w = w_0 - H^{-1}d. \qquad (2)$$

The form of the basic updating equation of the CG algorithm is the same as the general gradient algorithm, and is given by

$$w(k+1) = w(k) + \eta(k)\Delta w(k) \qquad (3)$$

where $\eta(k)$ is a time-varying learning parameter that may be updated using the following line search method:

$$\eta(k) = \min_{\eta}\{E(w(k) + \eta\Delta w(k)) : \eta \geq 0\} \qquad (4)$$

The conjugate condition for the incremental weight vector $\Delta w$ is designed as

$$(\Delta w(k))^T H(k)\Delta w(k+1) = 0 \qquad (5)$$

where the Hessian matrix $H(k)$ is calculated at the point $w(k)$. The updating for $w(k)$, in this case, is chosen as:

$$\Delta w(k) = -d(k) + \alpha(k-1)\Delta w(k-1) \qquad (6)$$

In order to satisfy the conjugate condition between the vectors $\Delta w(k)$ and $\Delta w(k+1)$, the update procedure for $\alpha(k)$ is expressed using the Fletcher-Reeves formulation, the Hestenes-Stiefel formulation and the Hestenes-Stiefel formulation given by equation 7, 8 and 9 respectively:

$$\alpha(k) = \frac{(d(k+1))^T d(k+1)}{(d(k))^T d(k)} \qquad (7)$$

$$\alpha(k) = \frac{(d(k+1))^T (d(k+1) - d(k))}{(d(k))^T d(k)} \qquad (8)$$

$$\alpha(k) = \frac{[d(k) - d(k-1)]^T d(k)}{(\Delta w(k-1))^T [d(k) - d(k-1)]} \qquad (9)$$

In fact, this algorithm exploits information about the direction of search for $\Delta w$ from the previous iteration in order to accelerate the convergence.

## 3. MIXTURE OF EXPERTS (ME)

MLPs have been used successfully for solving different regression and classification problems. However, for large problems the parameter space of MLPs becomes huge and hence in training phase it becomes computationally intractable. To tackle this problem, one can take advantage of the principle of "divide and conquer". According to the divide and conquer approach, one can solve a complex task by dividing it into simple tasks and then combining the solutions of them appropriately. A well-known method that works based on this principle is ME. This method is in the category of dynamic classifiers combining where the input signal is directly involved in the mechanism that integrates the output of the experts into an overall result [8]. For the first time ME was proposed in [3]. Their proposed model contains a population of simple linear classifiers (the experts). Outputs of the experts are mixed using a gating network. Technically the experts perform supervised learning, since for modeling the desired response; outputs of individual experts are combined. The experts are also self-organize to find a good partitioning of the input space and each expert model has its own subspace, and combination of all experts model the input space well. In [13, 14] this method was extended to the so-called "Hierarchical Mixture of Experts" (HME). In HME instead of single experts, mixtures of experts' models are used for each component. The ME has been studied for a wide range of research [15-18].

### 3.1. Mixture of Experts based on Gradient Decent Based (GDME)

In this version of Mixture of Experts, in order to improve the performance of the expert networks, MLPs are used instead of linear networks or experts. The learning algorithm is modified using an estimation of the posterior probability of the desired output by each expert. So, the gating and expert networks match and this improves the proposed model to select the best expert(s). The weights of MLPs expert networks are updated based on those estimations and the procedure is repeated.

As mentioned before, the convergence rate of GD learning algorithms is slow and hence it becomes unsuitable for large problems. To solve this problem we use Conjugate Gradient method in learning process of ME.

### 3.2. Mixture of Expert based on Conjugate Gradient (CGME)

In this section, we use Conjugate Gradient (CG) Methods for speeding up the learning phase of neural networks classifiers and gating network of ME. Each expert network is an MLP network with one hidden layer that computes an output $O_i$ as a function of the input vector, $x$ and weights of hidden and output layers and a sigmoid activation function. We assume that each expert specializes in a specific area of the input space. The gating network assigns a weight $g_i$ to each of the expert's output, $O_i$. The gating network determines $g_i$ as a function of the input vector $x$ and a set of parameters such as weights of its hidden and output layers. The activation function of the gating hidden layer is sigmoid, and in the output layer we use linear activation function. The softmax function was applied on the gating network outputs in order to make more diversity.

The $g_i$ can be interpreted as estimates of the prior probability that expert $i$ can generate the desired output $y$.

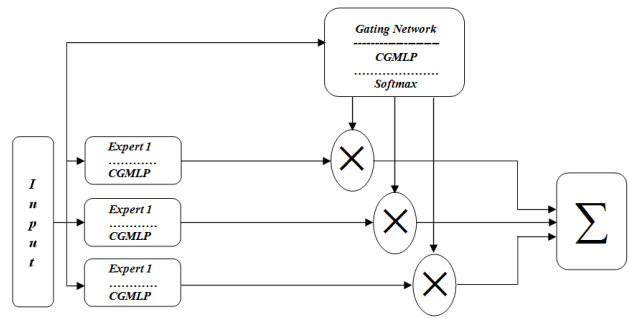

**Figure1** The block diagram of CGME

The experts compete to learn the training patterns and the gating network mediates the competition. Thus the gating network computes $O_g$, which is the output of the MLP layer of the gating network, then applies the softmax function to get:

$$g_i = \frac{\exp(O_{g\,i})}{\sum_{j=1}^{N} \exp(O_{g\,j})} \quad i = 1, ...., N \tag{10}$$

where $N$ is the number of expert networks, so $g_i$ s are nonnegative and sum up to 1. The final mixed output of the entire network is:

$$O_T = \sum_i O_i g_i \quad i = 1, ..., N \tag{11}$$

We train MLPs using the Conjugate Gradient algorithm. For each expert $i$ and the gating network, the weights are updated according to the following rules:

$$\Delta w_y(k) = \eta_e h_i [-(y - O_i) O_i (1 - O_i) + \alpha_y (k-1) \Delta w_y (k-1)] \tag{12}$$

$$\Delta w_h(k) = \eta_e h_i [-w_y^T (y - O_i)(O_i(1 - O_i)) O_{h_i} (1 - O_{h_i}) + \alpha_h (k-1) \Delta w_h (k-1)] \tag{13}$$

$$\Delta w_{yg}(k) = -\eta_g (h - g) O_g (1 - O_g) + \alpha_{yg} (k-1) \Delta w_{yg} (k-1) \tag{14}$$

$$\Delta w_{hg}(k) = -\eta_g w_{yg}^T (h - g) O_g (1 - O_g) O_{hg} (1 - O_{hg}) + \alpha_{hg} (k-1) \Delta w_{hg} (k-1) \tag{15}$$

where $\eta_e$ and $\eta_g$ are learning rates for the expert and the gating network, respectively. $w_h$ and $w_y$ are the weights of input to hidden and hidden to output layer, respectively, for experts $w_{hg}$ and $w_{yg}$ are the weights of input to hidden and hidden to output layer, respectively, for the gating network. $O_{hi}^T$ and $O_{hg}^T$ are the transpose of $O_{hi}$ and $O_{hg}$ which are the outputs of the hidden layer of expert and gating networks, respectively. $h_i$ is an estimation of the posterior probability that expert $i$ can generate for the desired output $y$:

$$h_i = \frac{g_i \exp(-\frac{1}{2}(y - O_i)^T (y - O_i))}{\sum_j g_j \exp(-\frac{1}{2}(y - O_j)^T (y - O_j))} \tag{16}$$

As pointed out in [19], in the network's learning process, the experts "compete" for explaining the input while the gate network rewards the winner of each competition using larger feedbacks. Thus, the gate divides the input space according to the performance of experts.

## 4. METAHEURISTIC ALGORITHM

Although CG converge faster than BP, but it remarkably depends on initial weights among neurons. If optimal weights are assigned to connections of neurons, it can be expected that the final performance increases. In order to tackle this problem, we applied a metaheuristic algorithm so called Modified Cuckoo Search.

During the 1950s and 1960s, computer scientists investigated the possibility of applying the concepts of evolution as an optimization tool for engineers and this gave birth to a subclass of gradient free methods called genetic algorithms (GA) [20]. Since then many other algorithms have been developed that have been inspired by nature, for example particle swarm optimization (PSO) [21], differential evolution (DE) [22] and, more recently, the cuckoo search (CS) [23]. These are heuristic techniques which make use of a large population of possible is design at each iteration. For each member of the population, the objective function is evaluated and a fitness is assigned. A set of rules is then used to move the population towards the optimum solution. Here, we used a new version of CS, the so called Modified Cuckoo Search (MCS) which was introduced by Walton et al [24]. They showed that MCS is more reliable and faster than CS, GA, DE and PSO [24].

## 4.1. Cuckoo Search (CS)

CS is a metaheuristic search algorithm which has been recently proposed by Yang and Deb [26]. The algorithm is inspired by the reproduction strategy of cuckoos. At the most basic level, cuckoos lay their eggs in the nests of other host birds, which may be of different species. The host bird may discover that the eggs are not it's own and either destroy the egg or abandon the nest all together. This has resulted in the evolution of cuckoo eggs which mimic the eggs of local host birds [25]. To apply this as an optimization tool, Yang and Deb [26] used three idealized rules:

- Each cuckoo lays one egg, which represents a set of solution co-ordinates, at a time and dumps it in a random nest;
- A fraction of the nests containing the best eggs, or solutions, will carry over to the next generation;
- The number of nests is fixed and there is a probability that a host can discover an alien egg, say $p_a \in [0,1]$. If this happens, the host can either discard the egg or the nest and these results in building a new nest in a new location.

---

**Algorithm 1.** Cuckoo Search (CS)

---

Initialize a population of $n$ host nests $x_i, i = 1, 2, ..., n$.

**for** all $xi$ **do**

    **1.1** Calculate fitness $Fi = f(xi)$

**end for**

**while** Number Objective Evaluations <Max Number Evaluations **do**

    Generate a cuckoo egg $(x_j)$ by taking a *Lévy* flight from random nest

    $F_j = f(x_j)$.

    Choose a random nest $I$

    **if** ( $F_j > F_i$ ) **then**

        $x_i \leftarrow x_j$

        $F_i \leftarrow F_j$

    **end if**

    Abandon a fraction $p_a$ of the worst nests Build new nests at new locations via *Lévy* flights to replace nests lost     Evaluate fitness of new nests and rank all solutions

**end while**

---

The steps involved in the CS are then derived from these rules and are shown in Algorithm 1 [23]. An important component of a CS is the use of *Lévy* flights for both local and global searching. The *Lévy* flight process, which has previously been used in search algorithms [27], is a random walk that is characterized by a series of instantaneous jumps chosen from a probability density function which has a power law tail Eq.7.

$$Lévy \sim u = t^{-\lambda} \quad 1 \leq \lambda \leq 3 \tag{17}$$

This process represents the optimum random search pattern and is frequently found in nature [28]. When generating a new egg in Algorithm 1, a *Lévy* flight is performed starting at the position of a randomly selected egg, if the objective function value at these new coordinates is better than another randomly selected egg then that egg is moved to this new position.

$$x_i^{(t+1)} = x_i^{(t)} + \alpha \oplus L\acute{e}vy(\lambda)$$

(18)

The scale of this random search is controlled by multiplying the generated *Lévy* flight by a step size $\alpha$. For example setting $\alpha = 0.1$ could be beneficial in problems of small domains, in the examples presented here $\alpha = 1$ is used in line with the work by Yang and Deb [26]. Yang and Deb [26] do not discuss boundary handling in their formulation, but an approach similar to PSO boundary handling [29] is adopted here. When a *Lévy* flight results in an egg location outside the bounds of the objective function, the fitness and position of the original egg are not changed. One of the advantages of CS over PSO is that only one parameter, the fraction of nests to abandon $p_a$, needs to be adjusted. Yang and Deb [26] found that the convergence rate was not strongly affected by the value and they suggested setting $p_a = 0.25$. The use of *Lévy* flights as the search method means that the CS can simultaneously find all optima in a design space and the method has been shown to perform well in comparison with PSO and GA [24].

## 4.2. Modified Cuckoo Search (MCS)

Given enough computation, the CS will always find the optimum [24] but, as the search relies entirely on random walks, a fast convergence cannot be guaranteed. In [25] two modifications to the method were made with the aim of increasing the convergence rate which leads to making the method more practical for a wider range of applications but without losing the attractive features of the original method. The first modification was made to the size of the Lévy flight step size $\alpha$. In CS, $\alpha$ is constant and the value $\alpha = 1$ is employed [26]. In the MCS, the value of $\alpha$ decreases as the number of generations increases. This is done for the same reasons that the inertia constant is reduced in the PSO [22], i.e. to encourage more localized searching as the individuals, or the eggs, get closer to the solution. An initial value of the Lévy flight step size $A = 1$ is chosen and, at each generation, a new Lévy flight step is calculated by using $\alpha = A/G$, where $G$ is the generation number. This exploratory search is only performed on the fraction of nests to be abandoned. The second modification is to add information exchange between the eggs in an attempt to speed up convergence to a minimum. In the CS, there is no information exchange between individuals and, essentially, the searches are performed independently. In the MCS, a fraction of the eggs with the best fitness are put into a group of top eggs. For each of the top eggs, a second egg in this group is picked at random and a new egg is then generated on the line connecting these two top eggs. The distance along this line at which the new egg is located is calculated, using the inverse of the golden ratio $\varphi = (1 + \sqrt{5})/2$, such that it is closer to the egg with the best fitness. In the case that both eggs have the same fitness, the new egg is generated at the midpoint. Whilst developing the method a random fraction was used in place of the golden ratio, it was found that the golden ratio showed significantly greater performance than a random fraction. There is a possibility that, in this step, the same egg is picked twice. In this case, a local Lévy flight search is performed from the randomly picked nest with step size $\alpha = A/G^2$. The steps involved in the modified cuckoo search are shown in detail in Algorithm 2. There are two parameters, the fraction of nests to be abandoned and the fraction of nests to make up the top nests, which need to be adjusted in the MCS. Through testing on benchmark problems, it was found that setting the fraction of nests to be abandoned to 0.75 and the fraction of nests placed in the top nests group to 0.25 yielded the best results over a variety of functions.

**Algorithm 2.** Modified Cuckoo Search (MCS)

$A \leftarrow MaxLévyStepSize$

$\varphi \leftarrow GoldenRatio$

Initialize a population of $n$ host nests $x_i$ $(i = 1, 2, ..., n)$

**for** all $x_i$ **do**

    Calculate fitness $F_i = f(x_i)$

**end for**

Generation Number $G \leftarrow 1$

**While** *Number Objective Evaluations < Max Number Evaluations* **do**

    $G \leftarrow G + 1$

    Sort nests by order of fitness

    **for** all nests to be abandoned **do**

        Current position $x_i$

        Calculate *Lévy* flight step size $\alpha \leftarrow A/\sqrt{G}$

        Perform *Lévy* flight from $x_i$ to generate new egg $x_k$

        $x_i \leftarrow x_k$

        $F_i \leftarrow f(x_i)$

    **end for**

    **for** all of the top nests **do**

        Current position $x_i$

        Pick another nest from the top nests at random $x_j$

        **if** $x_i = x_j$ **then**

            Calculate *Lévy* flight step size $\alpha \leftarrow A/G^2$

            Perform *Lévy* flight from $x_i$ to generate new egg $x_k$

            $F_k = f(x_k)$

            Choose a random nest $l$ from all nests

            **if** ( $F_k > F_l$ ) **do**

                $x_l \leftarrow x_k$

                $F_l \leftarrow F_k$

            **end if**

            **else**

                $dx = |x_i - x_j| / \varphi$

            Move distance $dx$ from the worst nest to the best nest to find $x_k$

            $F_k = f(x_k)$

            Choose a random nest $l$ from all nests

            **if** ( $F_k > F_l$ ) **then**

                $x_l \leftarrow x_k$

                $F_l \leftarrow F_k$

            **end if**

        **end if**

    **end for**

**end while**

# 5. EXPERIMENTAL RESULTS

In order to evaluate the performance of the proposed method, we performed three set of experiments. In the first experiment, we employed hybrid MCS and CGME method in function approximation problem. In second experiment, we tested this method on an artificial data set, and in third experiment we evaluated this method on seven UCI data sets [30].

## 5.1. Static function approximation

For the sake of comparison, the underlying function to be approximated is a three-input nonlinear function which is widely used to verify the efficiency of proposed algorithms [31]:

$$f = (1 + x^{0.5} + y^{-1} + z^{-1.5})^2 \qquad (19)$$

The training samples consist of 500 uniformly sampled three-input data from the input ranges $[1,6] \times [1,6] \times [1,6]$ and the corresponding target data. Other 250 testing samples are uniformly sampled from $[2,5] \times [2,5] \times [2,5]$, Table 1 are shown the error rate of MCS-CGME, GDME and CGME in the function approximation task.

**Table 1.** Comparison of MCS-CGME, CGME and GDME in function approximation task

|            | MSE  | Training set size | Test set size |
|------------|------|-------------------|---------------|
| GDME       | 0.26 | 500               | 250           |
| CGME       | 0.15 | 500               | 250           |
| MCS- CGME  | 0.11 | 500               | 250           |

## 5.2. Artificial Dataset

We consider a complex synthetic data set; this data set is generated using identical class of parametric distribution. In this data set, each class consists of three two-dimensional Gaussians:

$$C_i = \bigcup_{j=1}^{3} \left| \begin{matrix} N(\mu_{ij}, \delta_{ij}^2) \\ N(\mu'_{ij}, \delta_{ij}'^2) \end{matrix} \right. \quad \text{where } i = 1, 2, 3 \qquad (20)$$

The classes are parameterized as follows:

$$C_1 = \{ \left| \begin{matrix} N(6,1) \\ N(2,1) \end{matrix} \right., \left| \begin{matrix} N(14,1) \\ N(3,1) \end{matrix} \right., \left| \begin{matrix} N(18,1) \\ N(2,1) \end{matrix} \right. \} \qquad (21)$$

$$C_2 = \{ \left| \begin{matrix} N(5,1) \\ N(-1,1) \end{matrix} \right., \left| \begin{matrix} N(10.5,1) \\ N(3.5,1) \end{matrix} \right., \left| \begin{matrix} N(20,1) \\ N(0,1) \end{matrix} \right. \} \qquad (22)$$

$$C_3 = \{ \left| \begin{matrix} N(3,1) \\ N(2,1) \end{matrix} \right., \left| \begin{matrix} N(12,1) \\ N(6,1) \end{matrix} \right., \left| \begin{matrix} N(18,1) \\ N(-2,1) \end{matrix} \right. \}. \qquad (23)$$

Figure 2 shows our synthetic data set and class boundaries. Further, we summarize our experimental results on each method in Table 2.

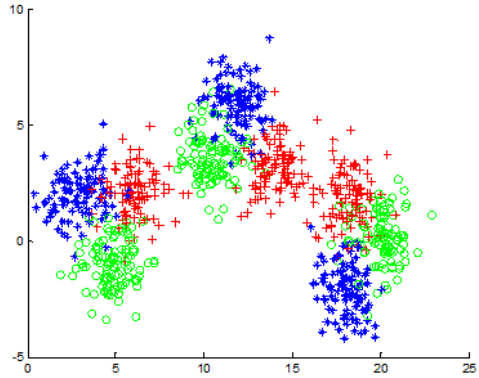

**Figure 2.** Our 3-class Artificial Data set.

**Table 2.** Performance of proposed method on Artificial Dataset

| Dataset | Performance (%) | | | | | | | |
|---|---|---|---|---|---|---|---|---|
| | **MLP** | | **GDME** | | **CGME** | | **MCS- CGME** | |
| | **Best** | **Average** | **Best** | **Average** | **Best** | **Average** | **Best** | **Average** |
| Artificial Dataset | 77.45 | 74.36 | 78.70 | 77.12 | 81.67 | 77.55 | 83.2 | 81.2 |

## 5.3. UCI data sets

Seven UCI data sets [30] are used in the experiments. Information of these data sets is shown in Table 3.

**Table 3.** UCI data set

| Data Set | Size | Attributes | Class |
|---|---|---|---|
| Sonar | 208 | 60 | 2 |
| Breast Cancer | 569 | 32 | 2 |
| Pima Indian Diabetes | 768 | 8 | 2 |
| Glass | 214 | 10 | 7 |
| Vehicle | 946 | 18 | 4 |
| Iris | 150 | 4 | 3 |
| Pen Digits | 10992 | 16 | 10 |

We compared our proposed model with GDME and CGME. Each of these ensemble models comprises five identical MLP networks initialized with (different) random weights, four of which are used as experts and the fifth as the gating network. The MLPs and the gating network consist of 5 and 15 sigmoid neurons in the hidden layer respectively. The classification performance is measured using 10-fold cross validation. Three ensemble models are trained using 100 epochs and the learning rate values of experts and gating networks were set to 0.1 and 0.15 respectively. The results are summarized in Table 4. To examine the superiority of our proposed combining method to stand-alone MLP, a single MLP network is used as baseline (c.f. Table 4). This MLP consists of one hidden layer with 25 neurons so the complexity of this MLP is similar to that of CGME and GDME.

**Table 4.** Recognition rate on USI datasets

| Dataset | Performance (%) | | | | | | | |
|---------|-----------------|---|---|---|---|---|---|---|
| | **MLP** | | **GDME** | | **CGME** | | **MCS- CGME** | |
| | **Best** | **Average** | **Best** | **Average** | **Best** | **Average** | **Best** | **Average** |
| Sonar | 94.62 | 93.65 | 97.60 | 95.67 | 99.33 | 98.11 | 100 | 99.23 |
| Breast | 97.60 | 94.45 | 98.07 | 96.84 | 99.12 | 98.59 | 99.57 | 99.12 |
| Pima | 70.96 | 68.13 | 78.26 | 77.76 | 83.77 | 79.22 | 85.28 | 82.36 |
| Glass | 61.68 | 60.56 | 65.42 | 64.48 | 79.07 | 67.20 | 81.13 | 70.85 |
| Vehicle | 77.42 | 76.60 | 81.56 | 80.58 | 81.91 | 80.65 | 82.56 | 81.23 |
| Iris | 92.35 | 89.68 | 97.18 | 95.43 | 99.46 | 98.06 | 100 | 99.57 |
| *Pen Digits | 85.05 | 84.02 | 90.70 | 86.12 | 92.50 | 88.75 | 94.36 | 93.67 |

([*]In our experiment 10% of this data set is used)

# 6. CONCLUSION

In this paper, an extended version of ME, MCS-CGME is introduced. In our proposed method, CG optimization technique is employed in BP learning algorithm of experts and gating network in ME model in order to tackle the problems associated with GD method. Moreover, MCS is applied in order to initial optimal weights for NN and as a result, convergence rate and performance of model are increased. Experimental results on regression and classification benchmark datasets demonstrate the supremacy of our proposed method in comparison with GDME and CGME.

**Authors**

**H. Salimi** was born in 1986 in Tehran, Iran. He has finished his Bachelor of Science in Computer Science from University of Tehran, under supervision of Dr. Noori. His research interests are Tensor decomposition, Statistical Learning, Computer Vision and Evolutionary Algorithms.

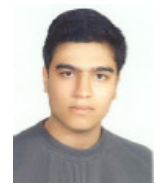

**D.Giveki** was born in 1984 in Boroujerd, Iran. He has graduated in Bachelor of Science in Computer Engineering and in Master of Science in Computer Science from University of Tehran. He is a PhD student in Computer Science in Saarland University. His research interests are image processing, machine learning and evolutionary algorithms.

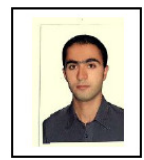

**M. A. Soltanshahi** was born in 1984 in Tehran, Iran. He has graduated in Bachelor of Science in Computer Science and in Master of Science in Computer Science from University of Tehran. He is the project manager and executive manager of center Jahad Daneshgahi University of Tehran. His research interests are Image Segmentation, Neural Networks and Parallel Algorithms.

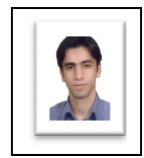

**J. Hatami** was born in 1987 in Masshhad, Iran. He has finished his Bachelor of Science in Computer Science from University of Tehran. His research interests are Image Processing, Statistical Learning, Computer Graphic, Human-computer Interaction and Evolutionary Algorithms.

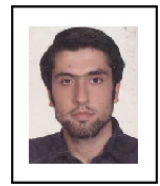